\documentclass[10pt,twocolumn,letterpaper]{article}

\usepackage[pagenumbers]{cvpr}
\usepackage{algorithm}
\usepackage{amssymb}
\usepackage{bbm}
\usepackage{algpseudocode}

\usepackage{amsthm}
\newtheorem{proposition}{Proposition}

\definecolor{cvprblue}{rgb}{0.21,0.49,0.74}
\usepackage[pagebackref,breaklinks,colorlinks,allcolors=cvprblue]{hyperref}

\def\confName{CVPR}
\def\confYear{2027}

\title{MPEG-Diffusion: Mixed-Resolution Video Generation}

\author{
Dennis Menn$^{1,2}$ \quad
Tianpei Gu$^{2}$ \quad
Guande He$^{2}$ \quad
Aryan V S$^{2}$ \quad
Yanke Song$^{2}$ \\
Nan Liu$^{2}$ \quad
Desai Xie$^{2}$ \quad
Xun Huang$^{2}$ \quad
Chenfeng Xu$^{1}$ \quad
Diana Marculescu$^{1}$ \\[0.5em]
$^{1}$UT Austin \qquad $^{2}$Roblox
}

\begin{document}
\twocolumn[{%
\renewcommand\twocolumn[1][]{#1}%
\maketitle
\vspace{-1.4em}
\begin{center}
  \centering
  \captionsetup{type=figure, skip=4pt}
  \includegraphics[width=\textwidth]{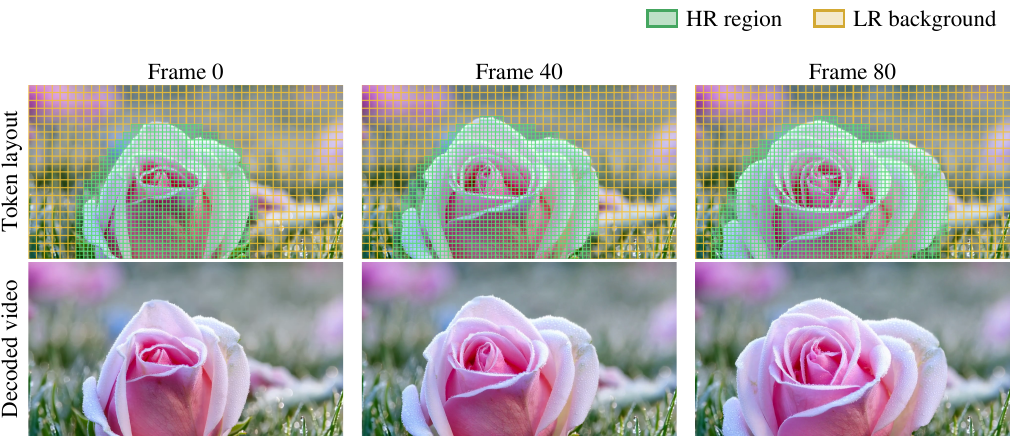}
  \captionof{figure}{\textbf{MPEG-Diffusion} generates the important region at high resolution and the rest at low resolution: on average \textbf{4.8$\boldsymbol{\times}$ faster (30\% token use)} at \textbf{near full-resolution quality}, with a \textbf{DiT self-selected high-res region}. Prompt: a close-up of a rose with morning droplets.}
  \label{fig:teaser}
\end{center}%
\vspace{0.4em}
}]

\begin{abstract}
Video diffusion transformers (DiTs) generate realistic videos, but their heavy computational cost limits broader
real-world applications. Mixed-resolution generation reduces this cost by generating the visually important region at
high resolution (HR) and the rest at low resolution (LR), yet existing methods suffer a severe quality drop on videos
and require knowing the HR region in advance. We show that these failures are not inherent to mixed resolution, and
propose \textbf{MPEG-Diffusion} to address their three root causes: (1) \emph{dual-resolution attention} to mitigate
the curse of mixed resolution, where each query reads the K/V of its own resolution, (2) a \emph{mixed-resolution
decoder} to blend the HR and LR latents before decoding and resolve distortion, and (3) a
\emph{self-selected HR region} using the DiT's knowledge to find the visually important region without requiring it in
advance. Processing only $30\%$ of full-resolution tokens, MPEG-Diffusion speeds up denoising by $4.8\times$ on Wan2.1
14B ($3.8\times$ on 1.3B), outperforming prior methods on nearly every metric. In the user study, we demonstrate the
high quality of MPEG-Diffusion, with raters preferring it over prior mixed-resolution methods in over $90\%$ of pairs
and rating it close to full resolution.
\end{abstract}
    
\section{Introduction}
\label{sec:intro}

\begin{figure*}[t]
  \centering
  \includegraphics[width=\textwidth]{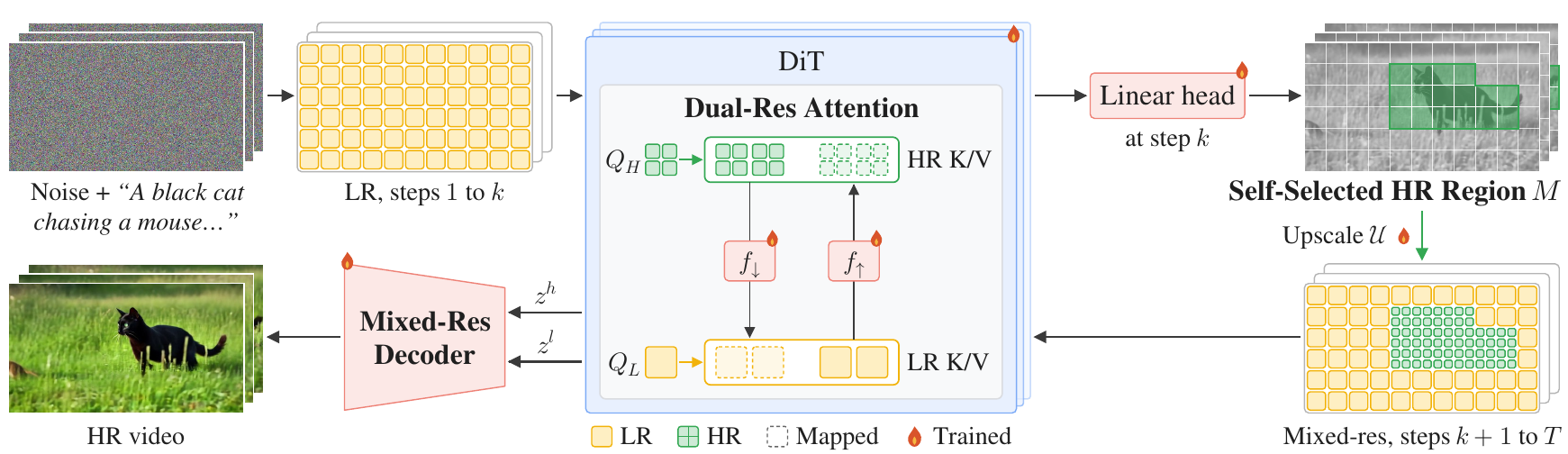}
  \caption{\textbf{Overview of MPEG-Diffusion.} The first $k$ of $T$ steps run at LR. At step $k$,
    a linear head selects the HR region $M$ (part~2), and the remaining steps denoise $M$ at HR and
    the rest at LR. In dual-resolution attention (part~1), each query reads the K/V of its own
    resolution, with missing entries (dashed) filled from the other resolution by linear maps. The
    mixed-resolution decoder (part~3) merges the HR and LR latents into one HR video. Flames mark trained parts.}
  \label{fig:overview}
\end{figure*}

Video diffusion transformers (DiTs)~\cite{wan2025} generate videos of striking quality, but attention scales
quadratically with the number of tokens as resolution and duration grow, limiting their broader real-world
applicability. Yet this compute is spent uniformly: every region of every frame is generated at the same resolution,
whether it shows a detailed face or a plain blue sky.

Video coding solved the same problem for transmission decades ago. Codecs such as MPEG and
H.264/AVC~\cite{wiegand2003overview} spend fewer bits on simple or predictable content and more on the regions that
matter, which saves a large share of the bandwidth at little perceptual cost. We ask whether a video generator can
do the same with \emph{compute}. Mixed-resolution generation~\cite{lipar,wu2025phase,foveateddiffusion} offers a natural
mechanism: the visually important region is generated at high resolution (HR) and the rest at low resolution (LR),
where one LR token replaces multiple HR tokens (\cref{fig:teaser}). This cuts the cost of every denoising step in proportion to the LR area.

In practice, however, mixed resolution is rarely used for video generation, for two reasons. First, prior
mixed-resolution methods suffer a sharp drop in quality, with blurry details and visible distortion along the
resolution boundary. Second, unlike a codec that compresses an existing video, a generator must know the HR region
in advance, when the intermediate frames are still too blurry for an off-the-shelf detector.

We argue that these failures are not inherent to mixed-resolution generation: we identify three root causes in
existing approaches~\cite{lipar,wu2025phase,foveateddiffusion} (\cref{sec:observations}) and propose
\emph{MPEG-Diffusion} (\cref{fig:overview}) to address each cause:
\begin{enumerate}
  \item \textbf{Curse of mixed resolution $\to$ Dual-resolution attention.} When HR and LR tokens share one set of
    keys and values (K/V), we prove that no single LR token can match the attention of the $n$ HR tokens it replaces
    (\cref{prop:curse}), even with fine-tuning. Instead, we propose \emph{dual-resolution attention}, where a linear
    map fills in the missing K/V from the other resolution (\cref{sec:obs_cache}), so each query reads the K/V of its
    own resolution, avoiding the quality drop caused by cross-resolution attention.
  \item \textbf{Resolution mismatch $\to$ Mixed-resolution decoder.} In the VAE decoder, each patch depends strongly on
    its neighboring patches, so decoding the LR and HR latents separately causes severe distortion at the resolution
    boundary, even on real videos (\cref{sec:obs_vae}); hence we propose a \emph{mixed-resolution decoder} to blend
    them in latent space before decoding.
  \item \textbf{External condition reliance $\to$ Self-selected HR region.} Existing methods require knowing the HR
    region in advance~\cite{lipar,foveateddiffusion}. Instead, leveraging the DiT's knowledge of the visually
    important subject across frames (\cref{fig:head_tracks}), we use a linear layer on the DiT features to predict the
    HR mask during denoising at negligible cost, enabling plain text-to-video generation.
\end{enumerate}

Processing only $30\%$ of full-resolution tokens, MPEG-Diffusion speeds up denoising by $3.8\times$ on Wan2.1 1.3B and
$4.8\times$ on 14B at near full-resolution quality, outperforming prior methods on nearly every metric at the same
token budget (\cref{tab:main}). In a blind user study, raters prefer ours over Phase-Aligned RoPE and Foveated
Diffusion ($98.8\%$ and $90.1\%$ win rates), with a $45.2\%$ win rate against full-resolution generation
(\cref{fig:human_eval}).

\section{Related work}\label{sec:related}

\paragraph{Efficient diffusion.} Along the token axis, token merging and pruning remove redundant tokens in the
intermediate layers of the transformer~\cite{bolya2022tome,bolya2023tome,li2024vidtome,fang2025sdtm,idm}, and sparse
attention skips unimportant query--key pairs in video DiTs~\cite{xi2025sparsevideogen,yang2025sparsevideogen2}. Along
the time axis, feature caching reuses intermediate features across denoising
steps~\cite{ma2024deepcache,liu2025teacache}, and distillation and consistency models reduce the number of
steps~\cite{salimans2022progressive,song2023consistency,yin2024dmd}. However, all of these methods still generate
every region at the same output resolution. We instead vary resolution across regions, orthogonal to these methods.

\paragraph{Mixed-resolution generation.} LIPAR~\cite{lipar} reuses temporal tokens that are deemed
redundant, whereas we focus on spatial redundancy. Phase-aligned RoPE~\cite{wu2025phase} and foveated
diffusion~\cite{foveateddiffusion} generate different regions at different resolutions, the latter around a known gaze
point. Concurrently, Level-of-Token Diffusion~\cite{lot2026} varies the token size across regions, following a layout
given in advance. While promising, on video generation their quality drops severely (\cref{tab:main}) and they require
the region in advance; we identify and address the three root causes (\cref{sec:observations}).

\paragraph{Localization from diffusion features.} Diffusion features carry rich semantic and spatial information. They
can match and track objects without supervision~\cite{tang2023dift,luo2023hyperfeatures,zhang2023tale}, support
segmentation~\cite{xu2023odise,zhao2023vpd,tian2024diffseg}, and their cross-attention maps localize the words of the
prompt~\cite{tang2023daam}. In video DiTs, attention also matches points across frames~\cite{nam2025difftrack}.
Building on this, we read the visually important HR region directly off the DiT with a single linear head
(\cref{sec:region}).

\section{Observations: why mixed resolution fails}\label{sec:observations}

We make three observations on why mixed resolution fails and how to fix it, each of which motivates one component
of our method (\cref{sec:method}).

\subsection{Observation 1: the curse of mixed resolution}\label{sec:obs_cache}

Whether an area is generated at HR or LR is a design choice: ideally, it should change only how much detail the area
has, and approximately maintain its effects on other regions. However, we show that under softmax attention, one LR
token cannot maintain the effects of multiple HR tokens when the two resolutions share one set of keys and values (a
\emph{shared K/V}). The mismatch comes from the number of tokens under softmax, not from the weights, so fine-tuning
alone cannot fully resolve it.

Consider an arbitrary softmax attention head in a transformer layer. Assume that we plan to replace $n$ high-resolution
tokens with keys $k_1,\dots,k_n$ and values $v_1,\dots,v_n$ by a low-resolution token with key $k_c$ and value
$v_c$. Let $\tilde k_1,\dots,\tilde k_m$ and $\tilde v_1,\dots,\tilde v_m$ be the keys and values of the other,
unchanged regions. Consider an arbitrary token with query $q$ (absorbing the $1/\sqrt{d}$ scale). Maintaining the
effects of the high-resolution tokens with one low-resolution token thus translates to (we use an equality for
simplicity, but the same conclusion also holds approximately):
\begin{equation}
  \frac{\sum_{i}e^{q\cdot k_i}v_i+\sum_{j}e^{q\cdot\tilde k_j}\tilde v_j}
       {\sum_{i}e^{q\cdot k_i}+\sum_{j}e^{q\cdot\tilde k_j}}
  =\frac{e^{q\cdot k_c}v_c+\sum_{j}e^{q\cdot\tilde k_j}\tilde v_j}
        {e^{q\cdot k_c}+\sum_{j}e^{q\cdot\tilde k_j}}.
  \label{eq:attn_match}
\end{equation}
Since high-to-low replacement does not control other regions, \cref{eq:attn_match} needs to hold for any combination
of $q$, $\{\tilde k_j\}_{j=1}^{m}$ and $\{\tilde v_j\}_{j=1}^{m}$. However, for $n\geq2$, even with a logit bias
$\beta$ as in proportional attention~\cite{bolya2022tome}, \cref{eq:attn_match} holds only in the trivial case where
all $n$ HR tokens are identical (\cref{prop:curse}). Replacing HR tokens with fewer LR tokens in a shared K/V is
therefore bound to affect other regions under softmax attention.

\begin{proposition}\label{prop:curse}
  Let $n\geq2$, $k_1,\dots,k_n,k_c\in\mathbb{R}^d$ and $\beta\in\mathbb{R}$, and add a bias $\beta$ to the logit of
  the LR token in \cref{eq:attn_match}, that is, replace $q\cdot k_c$ by $q\cdot k_c+\beta$. If \cref{eq:attn_match}
  holds for all $q$ in a non-empty open set $\Omega\subset\mathbb{R}^d$ and all $\{\tilde k_j\}$, $\{\tilde v_j\}$,
  then $k_1=\dots=k_n=k_c$, $\beta=\log n$ and $v_c=\frac{1}{n}\sum_i v_i$. The proof is in \cref{sec:supp_proof}.
\end{proposition}

\paragraph{From shared to per-resolution K/V.} A single video generative model can already generate good quality
videos at different resolutions when each video has a uniform resolution (so queries and K/V share the same
resolution). Mixed generation fails because LR queries attend to HR K/V or vice versa, breaking this match. Each query
should therefore read K/V at its own resolution, with a map filling in its missing positions from the other
resolution. Such a map is likely to exist because an LR token and the $n$ HR tokens of its area represent the same
content, differing only in fine detail.

\paragraph{A strong linear relation.} To test whether a simple linear map suffices, we encode 50 videos at both HR and
LR and run the base Wan2.1 1.3B DiT (without fine-tuning) on each resolution separately to collect the HR and LR K/V
at every block. Pairing each LR token with the $n$ HR tokens at the corresponding spatial-temporal position, we fit
one linear map per block and direction: the down map predicts the LR key (or value) from its $n$ corresponding HR keys
(or values), and the up map does the reverse. On 50 held-out videos, the fitted maps reach an average cosine
similarity of 0.96 and 0.85 for down keys and values, and 0.94 and 0.75 for up keys and values (\cref{fig:kv_cosine}
in the supplementary), indicating a strong linear relation between the two resolutions and making dual-resolution
attention feasible (\cref{sec:dual_cache}).

\subsection{Observation 2: VAE neighbor sensitivity}\label{sec:obs_vae}

\begin{figure}[t]
  \centering
  \includegraphics[width=\columnwidth]{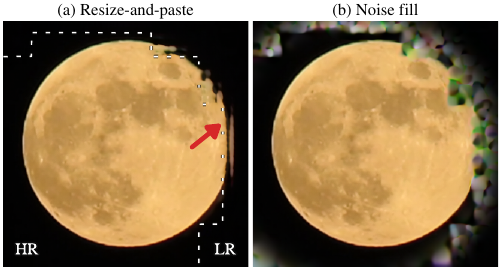}
  \caption{\textbf{The VAE decoder depends on neighboring latent patches.} (a) Resize-and-paste causes a seam along the resolution boundary (dashed; red arrow), even for latents encoded from a real video. (b) Noise fill spreads the error further.}
  \label{fig:vae}
\end{figure}

Due to the structural constraint of the VAE decoder, which requires a uniform spatial grid, a mixed-resolution latent
cannot be decoded directly because its HR and LR parts have different dimensions. A naive solution
(\emph{resize-and-paste}~\cite{foveateddiffusion}) fills each latent's missing region with the resized
other latent, decodes them separately, and pastes them together. However, we
find that the VAE decoder relies heavily on surrounding patches, so a misaligned neighbor corrupts the decoded patch
even when the patch itself is correct. To verify this without denoising error, we encode a real video at both HR and LR:
resize-and-paste still leaves a seam along the boundary (\cref{fig:vae}a), and filling
the missing region with noise spreads the artifacts further (\cref{fig:vae}b). This inspires our
\emph{mixed-resolution decoder} (\cref{sec:mixer}), which merges the HR and LR latents into a single unified latent
before decoding so every patch sees consistent neighbors.

\subsection{Observation 3: the DiT knows the important region}\label{sec:obs_head}

\begin{figure}[t]
  \centering
  \includegraphics[width=\columnwidth]{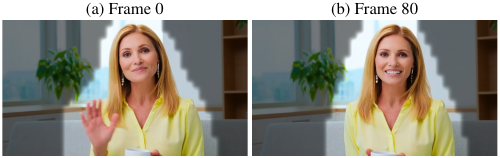}
  \caption{\textbf{The DiT knows where the subject is.} A linear head on the DiT features at an early denoising step predicts the HR region (bright), which tracks the subject from frame $0$ to $80$ while the background stays at LR (dimmed).}
  \label{fig:head_tracks}
\end{figure}

A major challenge of mixed resolution is knowing which region should be HR in advance. MPEG compresses a video that already
exists, so its motion vectors are known in advance. A generator, however, must choose its HR region before the video
exists, and early in denoising the predicted video is still too blurry for a detector to be reliable.

Surprisingly, we find that the DiT itself already knows which region is visually important. A single linear layer on the
DiT features captures the main subject, even as it moves (\cref{fig:head_tracks}). This matches prior work showing that
diffusion features can track objects~\cite{tang2023dift,nam2025difftrack}. This property lets us read the HR region
directly off the model without knowing it in advance, extending mixed-resolution generation to plain text-to-video
(\cref{sec:region}).

\section{MPEG-Diffusion}\label{sec:method}

Based on these observations, MPEG-Diffusion saves compute by generating the visually important regions of a video at
HR and the rest at LR (\cref{fig:overview}).

\paragraph{Inference.} Generation has four stages: (1) \textbf{LR generation:} the first $k$ of $T$ steps denoise the
whole video at LR; (2) \textbf{Region selection:} at step $k$, a linear head on the DiT features selects the HR region
$M$ (\cref{sec:region}); (3) \textbf{Mixed-resolution generation:} we upscale the selected LR region $M$ to HR with
$\mathcal{U}$ (by upscaling the clean LR prediction at step $k$ and re-noising it), and the remaining steps denoise
$M$ at HR and the rest at LR with dual-resolution attention (\cref{sec:dual_cache}); and (4) \textbf{Decoding:} the
mixed-resolution decoder merges the two latents into one HR video (\cref{sec:mixer}).

\paragraph{Training.} We fine-tune the DiT together with the K/V maps and a resolution embedding, one learned vector
each for HR and LR tokens, added to every token and initialized at zero. Each training video is encoded at both
resolutions and noised to the same level~\cite{foveateddiffusion}, and each latent learns to denoise only its own
region:
\begin{equation}
  \mathcal{L}_{\mathrm{DiT}} = \frac{\|M_{\uparrow}\odot(\hat{v}^h-v^h)\|^2}{\|M_{\uparrow}\|_1}
    + \frac{\|(1-M)\odot(\hat{v}^l-v^l)\|^2}{\|1-M\|_1},
  \label{eq:dit_loss}
\end{equation}
with $\hat{v}$ and $v$ the predicted and true velocities, and $M_{\uparrow}$ the mask on the HR grid. Here $M$ is the
25\% of tokens per latent frame with the highest perceptual label (\cref{sec:region}).

\subsection{Dual-resolution attention}\label{sec:dual_cache}

To avoid the curse of mixed resolution (\cref{prop:curse}), we propose \emph{dual-resolution attention}: each
attention layer keeps complete HR and LR sets of keys and values (K/V), and each query reads only the K/V of its own
resolution.

\paragraph{K/V mapping.} Since HR and LR tokens are only generated in their respective regions, we complete each
resolution's K/V from the other with linear maps $f(x)=xW$ (\cref{sec:obs_cache}), where $f_{\uparrow}$ maps one LR
token to $n$ HR tokens and $f_{\downarrow}$ maps $n$ HR tokens to one LR token ($n=4$ by default). With subscripts $H$
and $L$ for HR and LR tokens, the attention
is
\begin{align}
  O_H &= \mathrm{Attn}\Big(Q_H,\begin{bmatrix} K_H \\ f_{\uparrow}(K_L) \end{bmatrix},
         \begin{bmatrix} V_H \\ f_{\uparrow}(V_L) \end{bmatrix}\Big), \label{eq:hr_attn}\\
  O_L &= \mathrm{Attn}\Big(Q_L,\begin{bmatrix} K_L \\ f_{\downarrow}(K_H) \end{bmatrix},
         \begin{bmatrix} V_L \\ f_{\downarrow}(V_H) \end{bmatrix}\Big). \label{eq:lr_attn}
\end{align}
Each query thus sees the K/V of the whole video at its own resolution, as in single-resolution generation. Keys are
mapped before the rotary position embedding (RoPE)~\cite{su2024roformer}, so the maps act only on content and RoPE
uses each resolution's grid. The maps are fine-tuned jointly with the DiT.

We initialize each map by ridge regression on paired K/V, obtained by
running the base model on the same videos at both resolutions. For large models, full maps are costly in memory, so we
use a \emph{linear delta} variant instead, $W=A+UV^{\top}$, where $A$ is the exact average over the $n$ HR tokens
(down) or copy to them (up), and $UV^{\top}$ is a rank-$r$ correction.

\subsection{Mixed-resolution decoder}\label{sec:mixer}

\begin{figure}[t]
  \centering
  \includegraphics[width=\columnwidth]{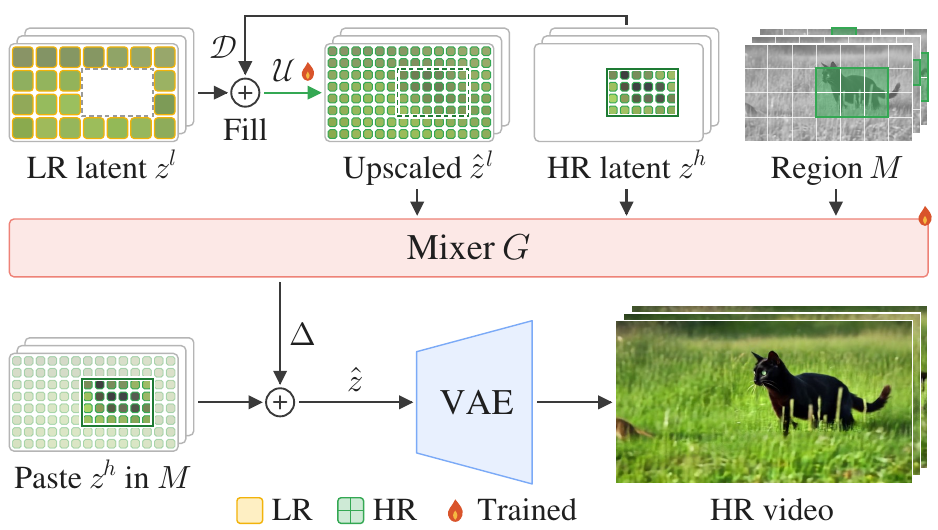}
 \caption{\textbf{Mixed-resolution decoder.} We merge the HR and LR latents before decoding. First, the LR latent
  is filled with the downscaled HR region ($\mathcal{D}$) and upscaled ($\mathcal{U}$), so patches near the boundary
  see less distorted neighbors. Then the mixer ($G$), guided by the mask, merges the LR and HR latents into one latent
  for the VAE.}
  \label{fig:mixer_design}

  \vspace{0.5em}
  \includegraphics[width=\columnwidth]{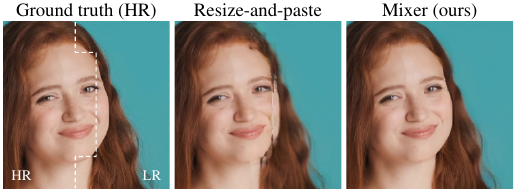}
  \caption{\textbf{The mixer removes the seam} across the HR/LR boundary (white dashed line). Left: ground-truth HR
    frame. Middle: resize-and-paste (\cref{sec:obs_vae}) leaves a seam across the face. Right: our mixed-resolution
    decoder keeps it intact.}
  \label{fig:mixer}
\end{figure}

After denoising, $z^h$ is generated only on the selected HR region $M$ and $z^l$ only on the remaining LR region. Due
to the resolution constraint of the VAE decoder (\cref{sec:obs_vae}), the two latents cannot be decoded together
directly, while decoding them separately without blending their information causes boundary seams. We therefore propose
a \emph{mixed-resolution decoder} that blends the HR and LR latents in latent space before decoding
(\cref{fig:mixer_design}).

\paragraph{Inference.} Since $z^l$ is missing the HR region $M$, we first fill $M$ with the downscaled HR latent
$\mathcal{D}(z^h)$ and upscale the completed LR latent to the HR grid with a learned latent upscaler
$\mathcal{U}$~\cite{wu2025phase} to match the size of $z^h$:
\begin{equation}
  \hat{z}^l=\mathcal{U}\big((1-M)\odot z^l+M\odot\mathcal{D}(z^h)\big).
  \label{eq:mix_up}
\end{equation}
Filling $M$ with $\mathcal{D}(z^h)$ provides cleaner boundary neighbors than nearest-neighbor resizing, giving a better
upscaled latent $\hat{z}^l$ before mixing. The mixer $G$ (a small 3D convolutional network) then takes $z^h$,
$\hat{z}^l$ and $M$, retains $z^h$ in the HR region, and predicts a residual $\Delta=G(z^h,\hat{z}^l,M)$ on the LR
region to remove the boundary seam:
\begin{equation}
  \hat{z}=M_{\uparrow}\odot z^h+(1-M_{\uparrow})\odot(\hat{z}^l+\Delta),
  \label{eq:mix_out}
\end{equation}
where $M_{\uparrow}$ is the mask on the HR grid. The VAE then decodes $\hat{z}$ into a seamless HR video
(\cref{fig:mixer}).

\paragraph{Training.} We train $G$ and $\mathcal{U}$ on real videos: given a training mask $M$ (\cref{sec:method}), we
keep only the HR latent in $M$ and the LR latent outside $M$, and train $\hat{z}$ to predict the ground-truth HR latent
(training recipe in \cref{sec:supp_mixer_loss}).

\subsection{Self-selected HR region}\label{sec:region}

\begin{figure}[t]
  \centering
  \includegraphics[width=\columnwidth]{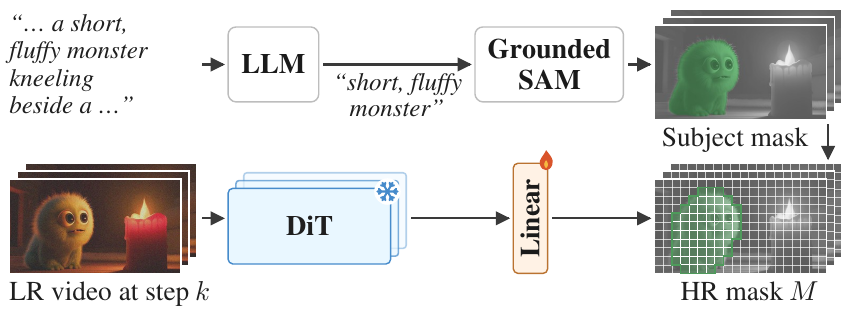}
  \caption{\textbf{Self-selected HR region.} \emph{Top:} an LLM identifies the main subject from the prompt, and Grounded SAM tracks it to produce subject masks. \emph{Bottom:} a linear head on the frozen DiT features at step $k$ learns to predict the HR mask $M$.}
  \label{fig:label_pipeline}
\end{figure}

To enable plain text-to-video generation, we leverage the DiT's ability to locate the visually important region
(\cref{sec:obs_head}) and read the HR region directly off its features with a linear head.

\paragraph{Inference.} At step $k$, from the output $h_b\in\mathbb{R}^{C}$ of a middle DiT block at LR token $b$, the
head scores each token as
\begin{equation}
  s_b=w^{\top}\mathrm{LN}(h_b),
  \label{eq:head}
\end{equation}
where $\mathrm{LN}$ is layer normalization and $w\in\mathbb{R}^{C}$ is the learned weight. We smooth the scores in
space and time, giving $\tilde s_b$, and threshold them relative to each video's own scores, with $m$ the median and
$p$ the 99th percentile of $\tilde s$:
\begin{equation}
  M_b = \mathbb{I}\big[\tilde s_b > m+\alpha\,(p-m)\big],
  \label{eq:threshold}
\end{equation}
with $\alpha$ a hyperparameter. If $p$ is below a fixed floor, the video has no clear subject and stays fully at LR.

\paragraph{Training.} We train the head on the frozen DiT and explore two ways to label visually important tokens: (1) a
self-supervised \emph{perceptual label}, defined as the per-patch LPIPS~\cite{zhang2018lpips} between the decoded LR
prediction at step $k$ and the ground-truth HR video (\cref{sec:supp_lpips}), which directly measures where LR loses
detail without external models, but can be distracted by high-frequency background texture; and (2) an \emph{object
label}, where an LLM~\cite{qwen25} identifies the main subject from the prompt and Grounded
SAM~\cite{liu2024grounding,ravi2024sam2} tracks it across frames (\cref{fig:label_pipeline}). We use the object label
at inference, as it better aligns with human attention, and train the head with binary cross-entropy loss.

\section{Experiments}\label{sec:settings}

\begin{figure*}[t]
  \centering
  \includegraphics[width=\textwidth]{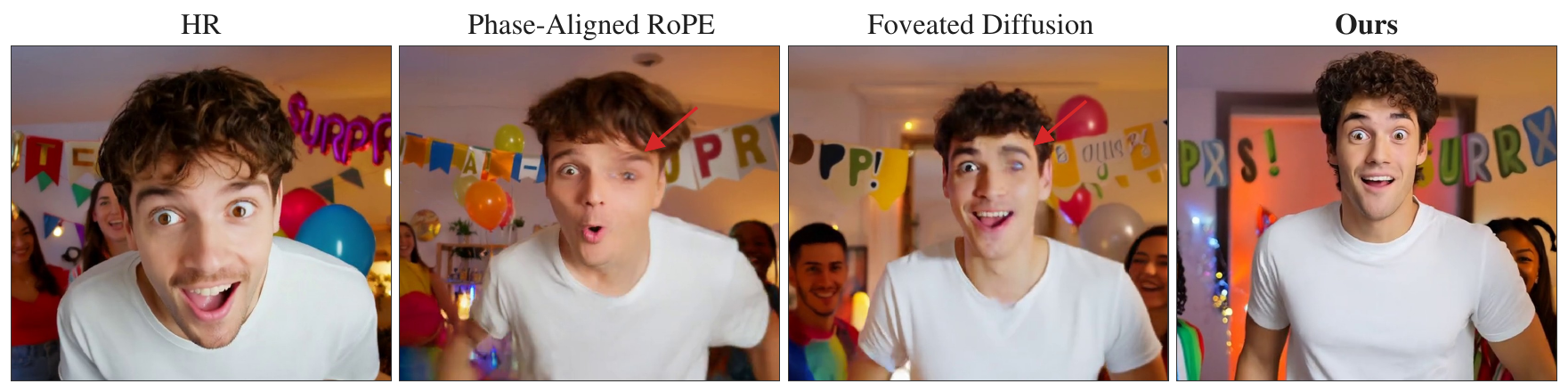}
  \caption{\textbf{Qualitative comparison on faces} (Wan2.1 14B, $k=25$, same seed). Prompt: a young man steps into
    a surprise party. Both baselines show clear artifacts where the HR and LR regions meet, with one eye sharp and the
    other ghosted (arrows). Ours shows no visible degradation and is close to HR (more examples in
    \cref{sec:supp_visuals}).}
  \label{fig:qualitative}
\end{figure*}

\paragraph{Setup.} We evaluate on both video and image generation. For video, we build on Wan2.1~\cite{wan2025} 1.3B
at 480p and 14B at 720p ($704{\times}1280$, so that each LR token covers $2{\times}2$ HR tokens), generating 81-frame
videos in 50 steps. The DiT is fully fine-tuned with the K/V maps for 1,250 steps on 64 NVIDIA B200 GPUs; the mixer
and upscaler are trained separately for 4k steps, and the region head on about 2,500 labeled clips. For images, we
apply our method to FLUX.2 [klein] 4B~\cite{flux2}, a DiT that, unlike Wan, puts the text and image tokens in one joint
attention, at $1024{\times}1024$.

\paragraph{Baselines.} We compare with token merging (ToMe~\cite{bolya2023tome}, IDM~\cite{idm}) and the prior
mixed-resolution methods, Phase-Aligned RoPE~\cite{wu2025phase} and Foveated Diffusion~\cite{foveateddiffusion}. Since
the gaze region is not available in text-to-video, we follow~\cite{foveateddiffusion} and run both with a circular HR
region moving along a predefined path at all steps, matching the token count and training data. HR and LR serve as
references.

\paragraph{Evaluation.} We generate one sample per prompt with a fixed seed. For video, we report
MANIQA~\cite{yang2022maniqa}, imaging quality~\cite{musiq} and DOVER~\cite{wu2023dover} on 1,003
MovieGen~\cite{moviegen2024} prompts and VBench Quality on 944 VBench~\cite{huang2024vbench} prompts, together with a
human study on 400 randomly selected video pairs from MovieGen (\cref{sec:main_results}). For images, we report
Q-Align~\cite{wu2024qalign} and patch FID~\cite{heusel2017fid} on 1,000 held-out prompts, together with a GPT-5.6
pairwise test (\cref{sec:supp_gpt}). Speed is the denoising time on one NVIDIA B200, relative to HR.

\subsection{Comparison with Acceleration Methods}
\label{sec:main_results}

\begin{table*}[t]
\centering
\caption{\textbf{Comparison at matched token budgets} ($k=40$ and $k=25$). The token ratio is the percentage of
  full-resolution tokens processed over all denoising steps. Speedup is measured in denoising time relative to HR.
  MANIQA, Imaging and DOVER are evaluated on the MovieGen prompt set~\cite{moviegen2024}, and VBench Quality on the
  VBench prompt set~\cite{huang2024vbench}.
  HR and LR are references. \textbf{Bold}: best among the token-matched methods.}
\label{tab:main}
\label{tab:budget25}
\footnotesize
\setlength{\tabcolsep}{4pt}
\begin{tabular}{@{}lcccccccccc@{}}
\toprule
& \multicolumn{5}{c}{Wan2.1 1.3B (480p)} & \multicolumn{5}{c}{Wan2.1 14B (720p)} \\
\cmidrule(lr){2-6} \cmidrule(l){7-11}
& Speedup & MANIQA & Imaging & DOVER & VBench Q.\ & Speedup & MANIQA & Imaging & DOVER & VBench Q.\ \\
Method & $\uparrow$ & $\uparrow$ & $\uparrow$ & $\uparrow$ & $\uparrow$ & $\uparrow$ & $\uparrow$ & $\uparrow$ & $\uparrow$ & $\uparrow$ \\
\midrule
HR ($100\%$ tokens) & 1.0$\times$ & 0.371 & 0.644 & 0.778 & 84.0 & 1.0$\times$ & 0.445 & 0.671 & 0.819 & 85.3 \\
LR ($25\%$ tokens)  & 5.8$\times$ & 0.244 & 0.586 & 0.492 & 80.3 & 7.2$\times$ & 0.318 & 0.645 & 0.768 & 84.4 \\
\midrule
\multicolumn{11}{@{}l}{\textit{$k=40$ ($30\%$ tokens)}} \\
ToMe~\cite{bolya2023tome}                   & 2.3$\times$ & 0.211 & 0.504 & 0.119 & 66.7 & 3.8$\times$ & 0.299 & 0.522 & 0.069 & 67.7 \\
IDM~\cite{idm}                              & 2.3$\times$ & 0.214 & 0.506 & 0.110 & 66.9 & 3.8$\times$ & 0.300 & 0.530 & 0.069 & 67.5 \\
Phase-Aligned RoPE~\cite{wu2025phase}       & 3.2$\times$ & 0.215 & 0.535 & 0.508 & 78.9 & \textbf{5.2$\times$} & 0.293 & 0.622 & 0.757 & 83.8 \\
Foveated Diffusion~\cite{foveateddiffusion} & 3.2$\times$ & 0.261 & 0.599 & 0.658 & 81.5 & \textbf{5.2$\times$} & 0.300 & 0.619 & 0.743 & \textbf{83.9} \\
MPEG-Diffusion                              & \textbf{3.8$\times$} & \textbf{0.345} & \textbf{0.631} & \textbf{0.762} & \textbf{81.7} & 4.8$\times$ & \textbf{0.397} & \textbf{0.644} & \textbf{0.794} & \textbf{83.9} \\
\midrule
\multicolumn{11}{@{}l}{\textit{$k=25$ ($38\%$ tokens)}} \\
Phase-Aligned RoPE~\cite{wu2025phase}       & 2.8$\times$ & 0.236 & 0.536 & 0.576 & 78.3 & \textbf{4.2$\times$} & 0.316 & 0.622 & 0.769 & 83.7 \\
Foveated Diffusion~\cite{foveateddiffusion} & 2.8$\times$ & 0.285 & 0.602 & 0.701 & 81.5 & \textbf{4.2$\times$} & 0.323 & 0.623 & 0.761 & 83.8 \\
MPEG-Diffusion                              & 2.8$\times$ & \textbf{0.355} & \textbf{0.635} & \textbf{0.769} & \textbf{81.6} & 3.3$\times$ & \textbf{0.402} & \textbf{0.649} & \textbf{0.803} & \textbf{84.1} \\
\bottomrule
\end{tabular}
\end{table*}

\paragraph{Qualitative.} We first compare generated videos with the prior mixed-resolution methods on both Wan2.1 14B
and 1.3B. In \cref{fig:qualitative}, both baselines show strong artifacts where the HR and LR regions meet, while our
method generates the subject sharply and without a seam. This holds across prompts and is even more pronounced when
viewed in video; we encourage the reader to view the videos in the supplementary material. Additional visual examples
with token layouts are provided in \cref{sec:supp_visuals}.

\begin{figure}[t]
  \centering
  \includegraphics[width=\columnwidth]{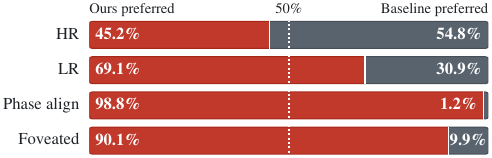}
  \caption{\textbf{Human evaluation} on Wan2.1 14B. Votes preferring ours (red) vs.\ baseline (gray),
  splitting draws 50/50.}
  \label{fig:human_eval}
\end{figure}

\paragraph{Machine metrics.} To quantitatively demonstrate the effectiveness of our method, we compare
MPEG-Diffusion with token merging and mixed-resolution methods on Wan2.1 1.3B and 14B at the same token ratio.
\textit{Quality.} At both $k=40$ (a $30\%$ token ratio) and $k=25$ ($38\%$ on
1.3B and $37\%$ on 14B, \cref{tab:main}), our method is the best on every quality metric for both models. The gap
is largest on MANIQA, which penalizes local distortions: prior mixed-resolution methods leave seams and artifacts
around the HR region, and ToMe~\cite{bolya2023tome} and IDM~\cite{idm} collapse into complete noise at this high
compression ratio. Our method also comes closest to HR (a DOVER gap of $0.016$ on 1.3B and $0.025$ on 14B at $k=40$),
preserving near full-resolution quality with only $30\%$ of the tokens needed.
While VBench Quality is insensitive to blur (barely separating HR and LR in the 14B case), detail-sensitive metrics (MANIQA, DOVER) and human raters (\cref{fig:human_eval}) show a clear gap over LR.

\textit{Speed.} On 1.3B, our method is the fastest at $k=40$ ($3.8\times$) and on par at $k=25$ ($2.8\times$). On
14B, it is slightly slower ($4.8\times$ vs.\ $5.2\times$ at $k=40$), because each HR token attends to the full HR K/V
rather than a shorter mixed-resolution sequence. However, with fewer tokens ($30\%$ vs.\ $37\%$), our method still beats both baselines in speed ($4.8\times$ vs.\ $4.2\times$) and on
every metric.

\paragraph{Human evaluation.} Metrics do not always match human judgment, so we also run a blind pairwise user study
on Wan2.1 14B. We compare ours with Phase-Aligned RoPE and Foveated Diffusion, all at a $37\%$ token ratio, HR and
LR, on 400 uncurated video pairs (100 MovieGen prompts $\times$ 4 baselines). Each of the 15 raters judges 200 pairs,
shown in random order with the two videos in random positions, and picks the one with better visual and temporal
quality, or a draw.

Following~\citet{alpaca_eval,genai_arena}, we split each draw equally between the two methods. Raters prefer ours over
Phase-Aligned RoPE ($98.8\%$), Foveated Diffusion ($90.1\%$) and LR ($69.1\%$) (\cref{fig:human_eval}). Against HR, a
draw is the most common answer ($44\%$), and HR is only slightly preferred ($54.8\%$ vs.\ $45.2\%$), though ours is
$3.3\times$ faster. Our method outperforms prior mixed-resolution methods and is close to
full-resolution quality at a fraction of the cost. Study details and win/draw/loss counts are in \cref{sec:supp_human}.

\begin{table}[t]
\centering
\caption{\textbf{Higher resolution} (Wan2.1 14B at 960p, $k=25$). Speedup is relative to HR. LR is generated at 240p.}
\label{tab:res960}
\footnotesize
\setlength{\tabcolsep}{3pt}
\resizebox{\columnwidth}{!}{%
\begin{tabular}{lccccc}
\toprule
Method & Speedup $\uparrow$ & MANIQA $\uparrow$ & Imaging $\uparrow$ & DOVER $\uparrow$ & VBench Q. $\uparrow$ \\
\midrule
\textit{HR} & \textit{1.0$\times$} & \textit{0.472} & \textit{0.667} & \textit{0.806} & \textit{85.3} \\
\textit{LR} & \textit{45.7$\times$} & \textit{0.270} & \textit{0.598} & \textit{0.592} & \textit{82.8} \\
\midrule
Foveated Diffusion & 4.7$\times$ & 0.354 & 0.604 & 0.766 & 81.6 \\
MPEG-Diffusion     & \textbf{4.8$\times$} & \textbf{0.367} & \textbf{0.633} & \textbf{0.789} & \textbf{82.9} \\
\bottomrule
\end{tabular}
}
\end{table}

\paragraph{Higher resolution.} To demonstrate the generalizability of MPEG-Diffusion, we also run Wan2.1 14B at 960p
with a 240p LR track, so each LR token stands for 16 HR tokens ($k=25$). Following the same evaluation setting as
\cref{tab:main}, we compare our method with Foveated Diffusion~\cite{foveateddiffusion} at the same token ratio. As
shown in \cref{tab:res960}, our method stays close to HR (DOVER $0.789$ vs.\ $0.806$) at $4.8\times$ the speed and
consistently beats Foveated Diffusion on every metric at the same speed, while LR degrades across all quality
metrics. The speedup also grows with resolution, from $3.3\times$ at 720p (\cref{tab:main}) to $4.8\times$ at 960p.

\subsection{Image generation}
\label{sec:image_gen}

\begin{figure}[t]
\centering
\captionof{table}{\textbf{Image generation on FLUX.2 [klein] 4B} ($1024{\times}1024$, 1,000 prompts). Speedup is per denoising
  step, relative to HR. FID: patch FID against the HR images. Ours pref.: rate at which GPT-5.6 prefers ours over the
  baseline ($>50\%$ favors ours). LR is a reference. \textbf{Bold}: best among the token-matched methods.}
\label{tab:flux}
\footnotesize
\setlength{\tabcolsep}{3pt}
\resizebox{\columnwidth}{!}{%
\begin{tabular}{@{}lccccc@{}}
\toprule
Method & Tokens (\%) $\downarrow$ & Speedup $\uparrow$ & Q-Align $\uparrow$ & FID $\downarrow$ & Ours pref.\ $\uparrow$ \\
\midrule
LR & 25 & 1.55$\times$ & 4.42 & 15.61 & 78.0\% \\
\midrule
Phase-Aligned RoPE & 33 & 1.11$\times$ & 4.34 & 14.59 & 83.8\% \\
Foveated Diffusion & 33 & 1.14$\times$ & 4.49 & 16.40 & 61.4\% \\
MPEG-Diffusion     & 33 & \textbf{1.40$\times$} & \textbf{4.60} & \textbf{12.97} & --- \\
\bottomrule
\end{tabular}
}

\vspace{0.5em}
\captionsetup{type=figure}
\includegraphics[width=\columnwidth]{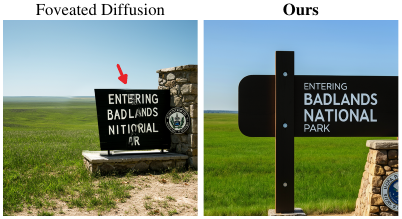}
\caption{\textbf{Text on FLUX.2 [klein] 4B} (same prompt, token budget and seed). Foveated Diffusion shows
  artifacts at the HR/LR seam (arrow), while ours generates the whole sign sharply (more examples in
  \cref{sec:supp_flux_visuals}).}
\label{fig:flux_sign}
\end{figure}

We further test MPEG-Diffusion on the text-to-image model FLUX.2 [klein]
4B~\cite{flux2} at $1024{\times}1024$. Prior mixed-resolution methods often show artifacts at
the seam between the HR and LR regions (\cref{fig:flux_sign}), while MPEG-Diffusion keeps the whole sign sharp and the
text correct (more examples in \cref{sec:supp_flux_visuals}).

To quantify this, we evaluate on 1,000 held-out prompts at a $33\%$ token ratio, following the default setting of
Foveated Diffusion~\cite{foveateddiffusion}. Our method is the best on every quality metric compared with the prior mixed-resolution methods and
LR (\cref{tab:flux}), with the lowest patch FID ($13.0$). We also perform a
blind pairwise test with GPT-5.6, which prefers our images over Foveated Diffusion, LR and Phase-Aligned RoPE in
$61.4\%$, $78.0\%$ and $83.8\%$ of the pairs.

Our method is also faster at the same token budget ($1.40\times$ vs.\ $1.14\times$ for Foveated Diffusion and
$1.11\times$ for Phase-Aligned RoPE). It runs the first $k$ steps purely at LR, whereas both baselines run mixed
resolution at all steps and split every attention layer into two calls, which incurs large overhead when each
image has only 4,096 tokens.

\subsection{Ablation studies}
\label{sec:ablation}

\begin{figure}[t]
  \centering
  \includegraphics[width=\columnwidth]{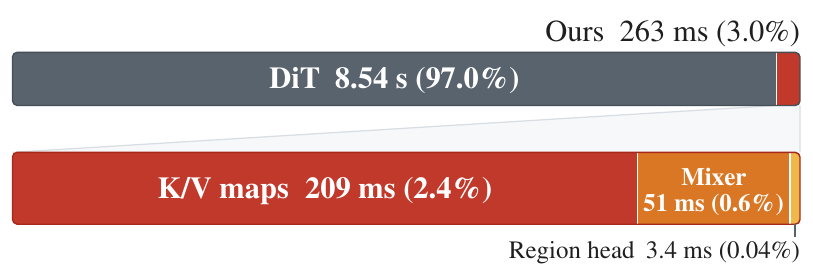}
  \caption{\textbf{Runtime breakdown on Wan2.1 1.3B.} Our components add only $3.0\%$ overhead, where K/V maps account for $2.4\%$.}
  \label{fig:overhead}

  \vspace{0.5em}
  \includegraphics[width=\columnwidth]{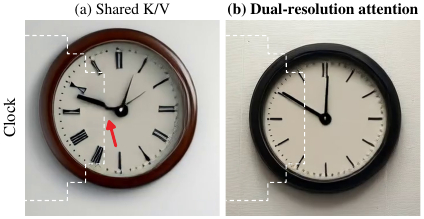}
  \caption{\textbf{Misalignment across the HR/LR edge} (white dashed circle). A shared K/V distorts the clock hand
    crossing the edge (red arrow), whereas dual-resolution attention keeps it straight.}
  \label{fig:kv_misalignment}
\end{figure}

\paragraph{Runtime overhead.} We profile the runtime of MPEG-Diffusion on Wan2.1 1.3B (480p, 81 frames, 50 steps, one
GPU, batch size 1) at a $30\%$ token ratio, averaging nine runs after one warm-up run. The MPEG components add only $3.0\%$
to the generation time: the K/V maps take $2.4\%$, the mixer $0.6\%$ and the region head
$0.04\%$, as shown in \cref{fig:overhead}.

\begin{figure}[t]
  \centering
  \makebox[\columnwidth][c]{%
    \begin{tabular}{@{}c@{\hspace{0.01\columnwidth}}c@{}}
      \includegraphics[trim=4 2 8 6,clip,width=0.515\columnwidth,height=0.37\columnwidth]{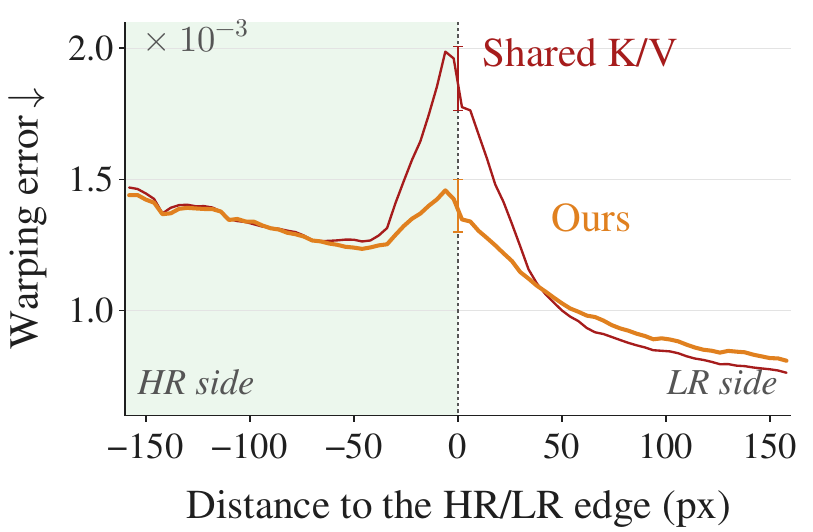} &
      \includegraphics[trim=4 2 4 6,clip,width=0.515\columnwidth,height=0.37\columnwidth]{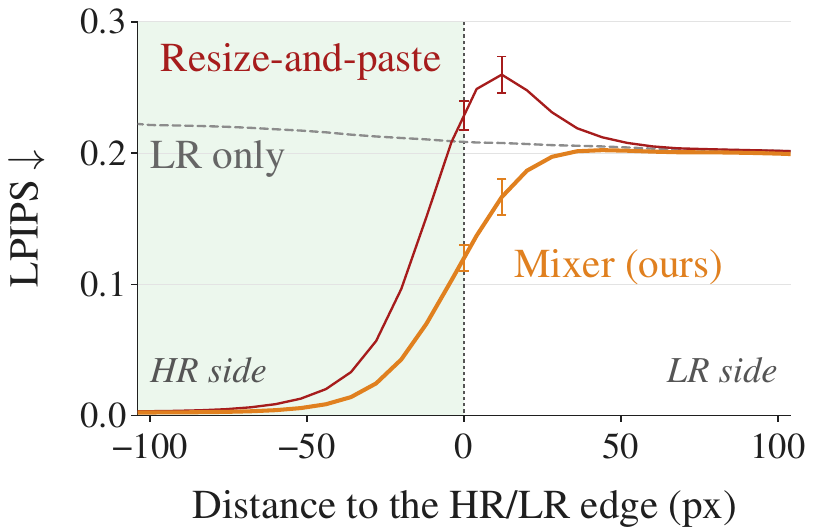} \\
      \parbox[t]{0.50\columnwidth}{\centering\footnotesize (a) Dual-resolution attention: warping error, 1,003 prompts} &
      \parbox[t]{0.50\columnwidth}{\centering\footnotesize (b) Mixed-resolution decoder: LPIPS, 100 real clips} \\
    \end{tabular}%
  }
  \caption{\textbf{Error profile across the HR/LR edge} (negative distance: HR side; shaded: $95\%$ CI). (a)
    Dual-resolution attention suppresses the warping-error spike of the shared K/V at the edge ($+25\%$ vs.\ $+70\%$).
    (b) Our mixer removes the LPIPS spike of resize-and-paste just outside the edge.}
  \label{fig:edge_profiles}
\end{figure}

\paragraph{Dual-resolution attention.} In \cref{sec:obs_cache}, we show that with a shared K/V, one LR token cannot
match the attention of the $n$ HR tokens it replaces, which motivates dual-resolution attention. We compare it with a
shared K/V with adjusted RoPE, as in Foveated Diffusion~\cite{foveateddiffusion} and Phase-Aligned
RoPE~\cite{wu2025phase}, both on Wan2.1 1.3B, trained for 5,000 steps, with all else fixed, and evaluated on the
1,003 MovieGen prompts (details in \cref{sec:supp_kv_ablation}).

In \cref{fig:kv_misalignment}, the resolution mismatch in the shared K/V bends the clock hand crossing the HR/LR edge,
whereas dual-resolution attention keeps it straight. To quantify this phenomenon, we measure temporal
consistency near the edge with temporal LPIPS~\cite{chu2020tecogan} and the warping error~\cite{lai2018blind}. As the
HR region moves, the seam moves with it, so the distortion at the seam changes over time and lowers temporal
consistency.

At the seam between HR and LR, dual-resolution attention lowers temporal LPIPS from $4.59$ to $4.14\times 10^{-2}$ and
the warping error from $2.61$ to $1.96\times 10^{-3}$ (\cref{tab:kv_ablation} in the supplementary), indicating less
distortion. \Cref{fig:edge_profiles}a further confirms this: at the seam, the
warping error is much lower with dual-resolution attention, while far from the seam the two models are the same.

\paragraph{Mixed-resolution decoder.} In \cref{sec:obs_vae}, we observe that the VAE decoder depends on neighboring
patches, so resizing and pasting the two decodes leaves a seam.

To test whether the mixed-resolution decoder removes this seam, we use 100 real held-out videos. We encode each video
at 480p and at 240p, and build a mixed latent as in generation: the 480p latent inside a random moving circle and the
240p latent outside. We measure the LPIPS of each decoding method on the mixed-resolution latent against the 480p
video in pixel space, comparing the mixer with resize-and-paste and LR only.

In \cref{fig:edge_profiles}b, we find that near the HR/LR edge, resize-and-paste shows a surge in LPIPS, indicating a
distortion relative to the 480p video. In contrast, the mixer removes the distortion, lowering LPIPS near the edge
from $0.19$ to $0.12$.

\paragraph{Self-selected region.} In \cref{sec:obs_head}, we observe that DiT features can locate the visually important
region. To evaluate our linear head quantitatively, we compare the HR mask $M$ it predicts at step $k$ against the
main-subject mask labeled by the LLM and Grounded SAM (\cref{sec:region}) across the 1,003 generated MovieGen videos.
The predicted region achieves an mIoU of $0.59$ on Wan2.1 1.3B and $0.67$ on 14B, showing that the linear head
accurately locates and tracks the main subject during denoising.

\section{Conclusion}
\label{sec:conclusion}
We studied why mixed-resolution generation comes with a severe drop in quality. We found three root causes and
addressed each with one component of MPEG-Diffusion: (1) dual-resolution attention to mitigate the curse of mixed
resolution, (2) the mixed-resolution decoder to blend the HR and LR latents before decoding, and (3) a linear layer
using the DiT's knowledge to find the important region without requiring it in advance. Together, MPEG-Diffusion
achieves up to a $4.8\times$ speedup, and in our human study raters prefer it over prior mixed-resolution methods in
over $90\%$ of the pairs. We hope our work opens the way to mixed-resolution generation.

\paragraph{Limitations and future work.} Dual-resolution attention assumes a linear relation between the HR and LR
tokens of an area. This relation is strong in the base model (\cref{sec:obs_cache}), but it remains a structural
constraint. This may still lead to a slight degradation of the quality. Additionally, the difference between the
blurry LR region and the sharp HR details is sometimes obvious, and further work is needed to blend them more
smoothly. Finally, the self-selected region is learned from subject labels given by an LLM and Grounded SAM, so we
suggest end-to-end fine-tuning as future work.

\section*{LLM Usage Statement}
\label{sec:llm_usage}
LLMs are used in two parts of the method and evaluation, both described in the paper: Qwen2.5~\cite{qwen25}
identifies the main subject from the prompt to build the object labels for the region head (\cref{sec:region}), and
GPT-5.6 serves as a blind pairwise judge of image quality (\cref{sec:image_gen}). We also used an LLM-based assistant to edit the text for grammar and clarity. The authors checked all text and results and take full responsibility for them.

{
    \small
    \bibliographystyle{ieeenat_fullname}
    \bibliography{main}
}

\clearpage
\setcounter{page}{1}
\appendix

\twocolumn[{%
\renewcommand\twocolumn[1][]{#1}%
{\centering\Large\textbf{\thetitle}\\
\vspace{0.35em}Supplementary Material \\
\vspace{0.55em}}
\section{Additional video qualitative results}\label{sec:supp_visuals}
\Cref{fig:supp_bird,fig:supp_glass,fig:supp_hug,fig:supp_hiker,fig:supp_sunglasses} show additional visual examples
generated by MPEG-Diffusion on Wan2.1 14B ($k=25$), pairing each clip's token layout with its decoded video as in
\cref{fig:teaser}.
\vspace{0.15em}
\begin{center}
  \centering
  \captionsetup{type=figure, skip=1pt}
  \includegraphics[width=0.83\textwidth]{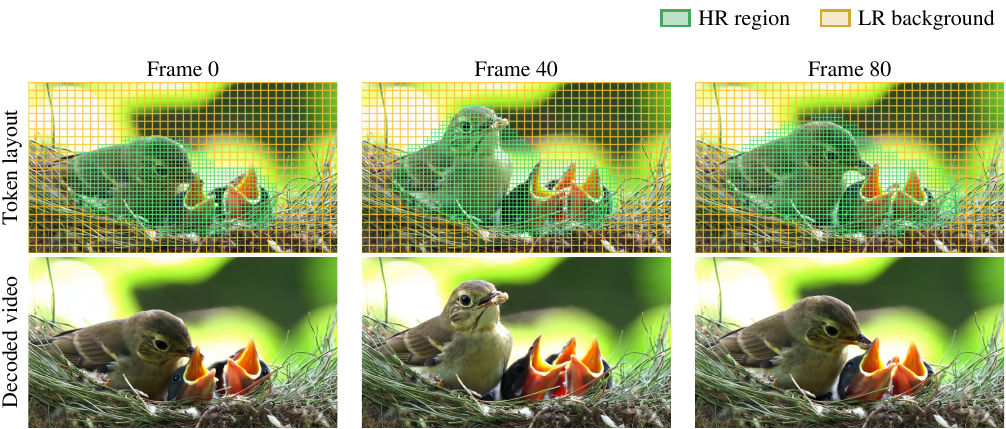}
  \captionof{figure}{\textbf{Mother bird feeding chicks} (\textbf{3.1$\boldsymbol{\times}$ faster}, \textbf{38.7\%} token use). Prompt: a mother bird feeding her chicks in a nest.}
  \label{fig:supp_bird}
  \vspace{0.08em}
  \includegraphics[width=0.83\textwidth]{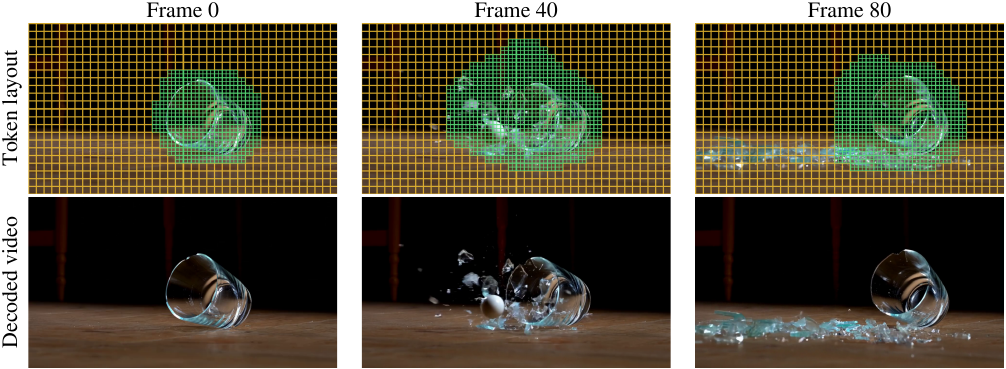}
  \captionof{figure}{\textbf{Marble through a glass} (\textbf{3.4$\boldsymbol{\times}$ faster}, \textbf{36.1\%} token use). Prompt: a marble shattering a glass cup on a wooden table.}
  \label{fig:supp_glass}
  \vspace{0.08em}
  \includegraphics[width=0.83\textwidth]{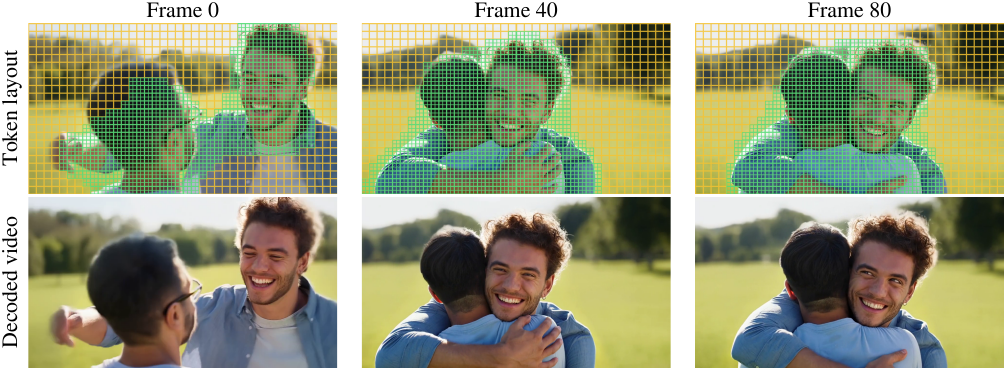}
  \captionof{figure}{\textbf{Friends hugging} (\textbf{2.4$\boldsymbol{\times}$ faster}, \textbf{45.0\%} token use). Prompt: two friends hugging outdoors.}
  \label{fig:supp_hug}
\end{center}
}]

\twocolumn[{%
\renewcommand\twocolumn[1][]{#1}%
\begin{center}
  \centering
  \captionsetup{type=figure, skip=2pt}
  \includegraphics[width=0.92\textwidth]{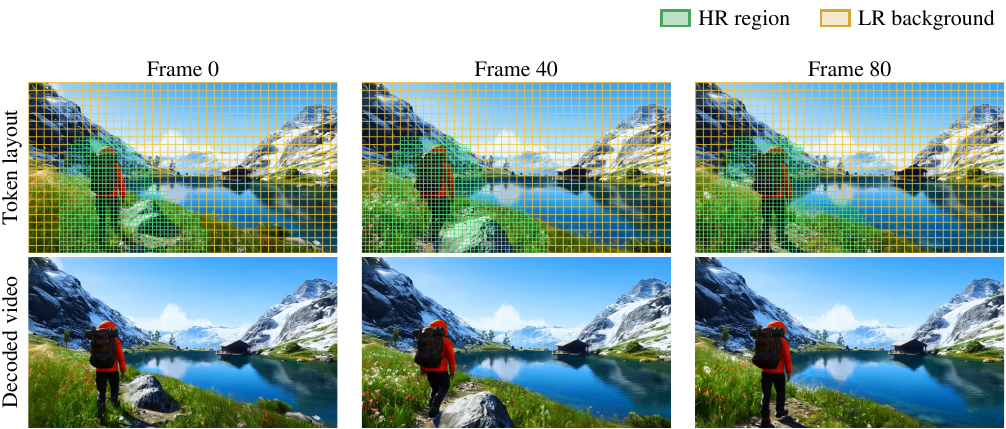}
  \captionof{figure}{\textbf{Hiker by an alpine lake} (\textbf{3.4$\boldsymbol{\times}$ faster}, \textbf{36.1\%} token use). Prompt: a hiker walking beside an alpine mountain lake.}
  \label{fig:supp_hiker}
  \vspace{0.6em}
  \includegraphics[width=0.92\textwidth]{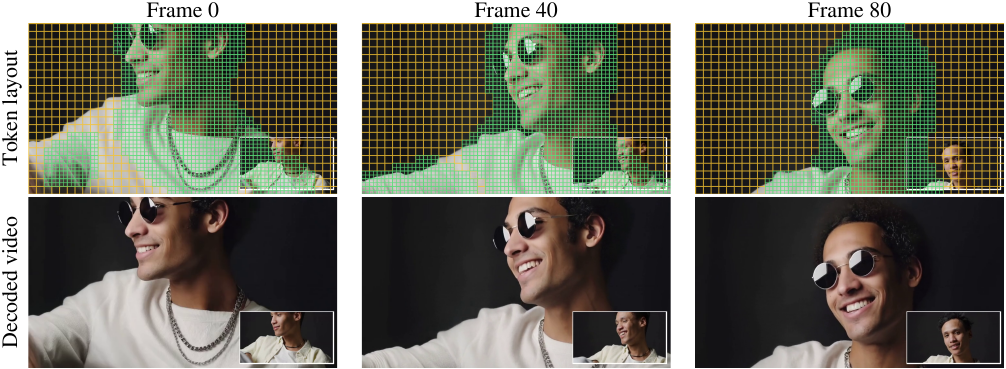}
  \captionof{figure}{\textbf{Sunglasses and leather jacket} (\textbf{4.0$\boldsymbol{\times}$ faster}, \textbf{48.2\%} token use, input video in the corner). Prompt: a young man wearing black sunglasses and a leather jacket.}
  \label{fig:supp_sunglasses}
\end{center}
\vspace{0.5em}
}]
\clearpage

\twocolumn[{%
\renewcommand\twocolumn[1][]{#1}%
\section{Additional image qualitative results}\label{sec:supp_flux_visuals}
\Cref{fig:supp_flux1,fig:supp_flux2} show additional text-to-image examples generated by MPEG-Diffusion on FLUX.2
[klein] 4B ($1024{\times}1024$), pairing each image's token layout with its decoded image.
\vspace{0.15em}
\begin{center}
  \centering
  \captionsetup{type=figure, skip=1pt}
  \includegraphics[width=0.77\textwidth]{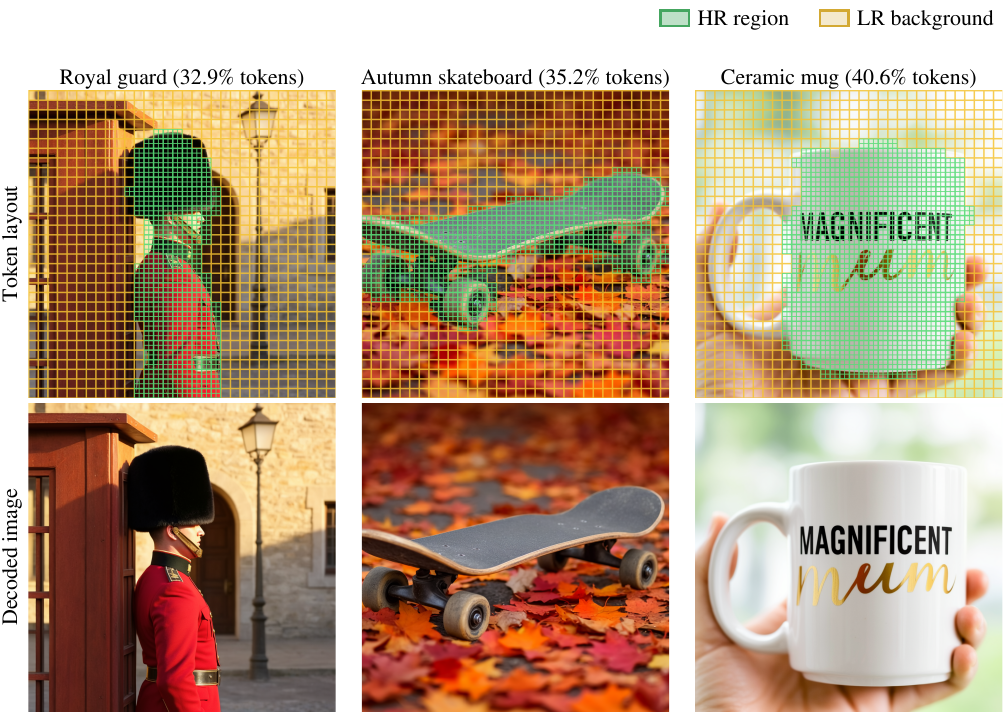}
  \captionof{figure}{\textbf{Text-to-image generation on FLUX.2 [klein] 4B} ($1024{\times}1024$, \textbf{1.40$\boldsymbol{\times}$ faster} on average). Left: a royal guard in a red uniform and bearskin hat in a sentry box. Middle: a skateboard on colorful autumn leaves. Right: a hand holding a ceramic mug reading MAGNIFICENT Mum.}
  \label{fig:supp_flux1}
  \vspace{0.15em}
  \includegraphics[width=0.77\textwidth]{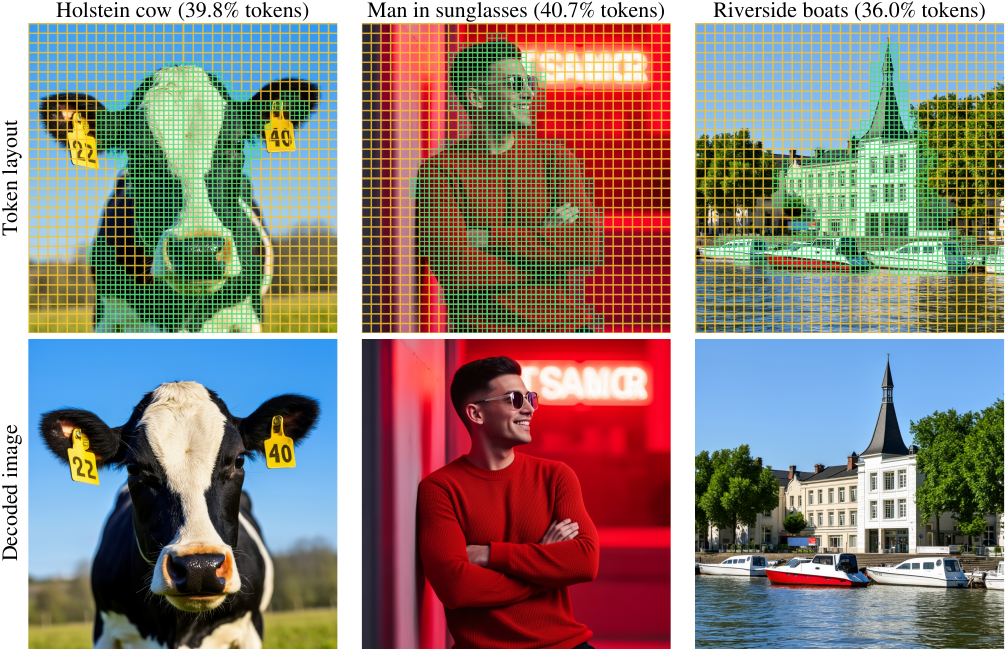}
  \captionof{figure}{\textbf{Additional text-to-image results on FLUX.2 [klein] 4B} ($1024{\times}1024$). Left: a Holstein cow in a field under a blue sky. Middle: a young man in sunglasses leaning against a neon-lit wall. Right: boats anchored near a historic riverside building.}
  \label{fig:supp_flux2}
\end{center}
}]
\clearpage

\section{Proof of \cref{prop:curse}}\label{sec:supp_proof}

\begin{proof}
  Fix $q$ and $\{\tilde k_j\}$. Both sides of \cref{eq:attn_match} are affine in
  $R=\sum_j e^{q\cdot\tilde k_j}\tilde v_j$, which takes any value as the $\tilde v_j$ vary. Matching the coefficients
  of $R$ gives $e^{q\cdot k_c+\beta}=\sum_i e^{q\cdot k_i}$, and matching the rest gives
  $e^{q\cdot k_c+\beta}v_c=\sum_i e^{q\cdot k_i}v_i$. Taking logs of the first,
  \begin{equation}
    \underbrace{q\cdot k_c+\beta}_{g(q)}=\underbrace{\log\sum_{i=1}^{n}e^{q\cdot k_i}}_{f(q)}
    \quad\text{on }\Omega.
    \label{eq:match}
  \end{equation}
  Let $p_i(q)=e^{q\cdot k_i}/\sum_j e^{q\cdot k_j}>0$ be the
  softmax weights. Then $\nabla f(q)=\sum_i p_i k_i=:\bar{k}(q)$ and
  \begin{equation*}
    \nabla^2 f(q)=\sum_{i=1}^{n}p_i\,(k_i-\bar{k})(k_i-\bar{k})^{\top}.
  \end{equation*}
  The left side of \cref{eq:match} is affine in $q$, so this Hessian is zero on $\Omega$, and so is its trace
  $\sum_i p_i\|k_i-\bar{k}\|^2$. Since every $p_i>0$, all $n$ keys equal $\bar{k}$; call them $k$. Then
  $f(q)=q\cdot k+\log n$, and \cref{eq:match} becomes $q\cdot(k_c-k)=\log n-\beta$ on $\Omega$. An affine function
  that is constant on an open set has zero slope, so $k_c=k$ and $\beta=\log n$. Finally, the second condition
  becomes $n\,e^{q\cdot k}v_c=e^{q\cdot k}\sum_i v_i$, so $v_c=\frac{1}{n}\sum_i v_i$.
\end{proof}

\section{Linear relation between HR and LR keys and values}\label{sec:supp_kv}

\begin{figure}[h!]
  \centering
  \includegraphics[width=0.82\columnwidth]{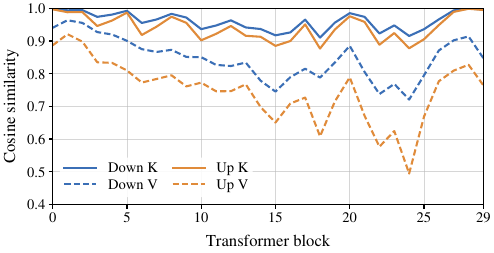}
  \caption{\textbf{A linear map predicts the K/V of the other resolution.} Cosine similarity between the mapped and
    native keys (K) and values (V) in each transformer block of the base Wan2.1 1.3B model (no fine-tuning) at $t=0$,
    on 50 held-out clips (about 1.6M HR and 0.4M LR tokens). Down: HR $\to$ LR; Up: LR $\to$ HR.}
  \label{fig:kv_cosine}
\end{figure}

\cref{fig:kv_cosine} shows, for each transformer block, how well the linear maps of \cref{sec:obs_cache} predict the
keys and values of the other resolution.

\section{Human evaluation details}\label{sec:supp_human}

This section details the human study of \cref{sec:main_results}: how it is run, what raters
see, and the full vote counts.

\begin{figure}[h!]
  \centering
  \includegraphics[width=\columnwidth]{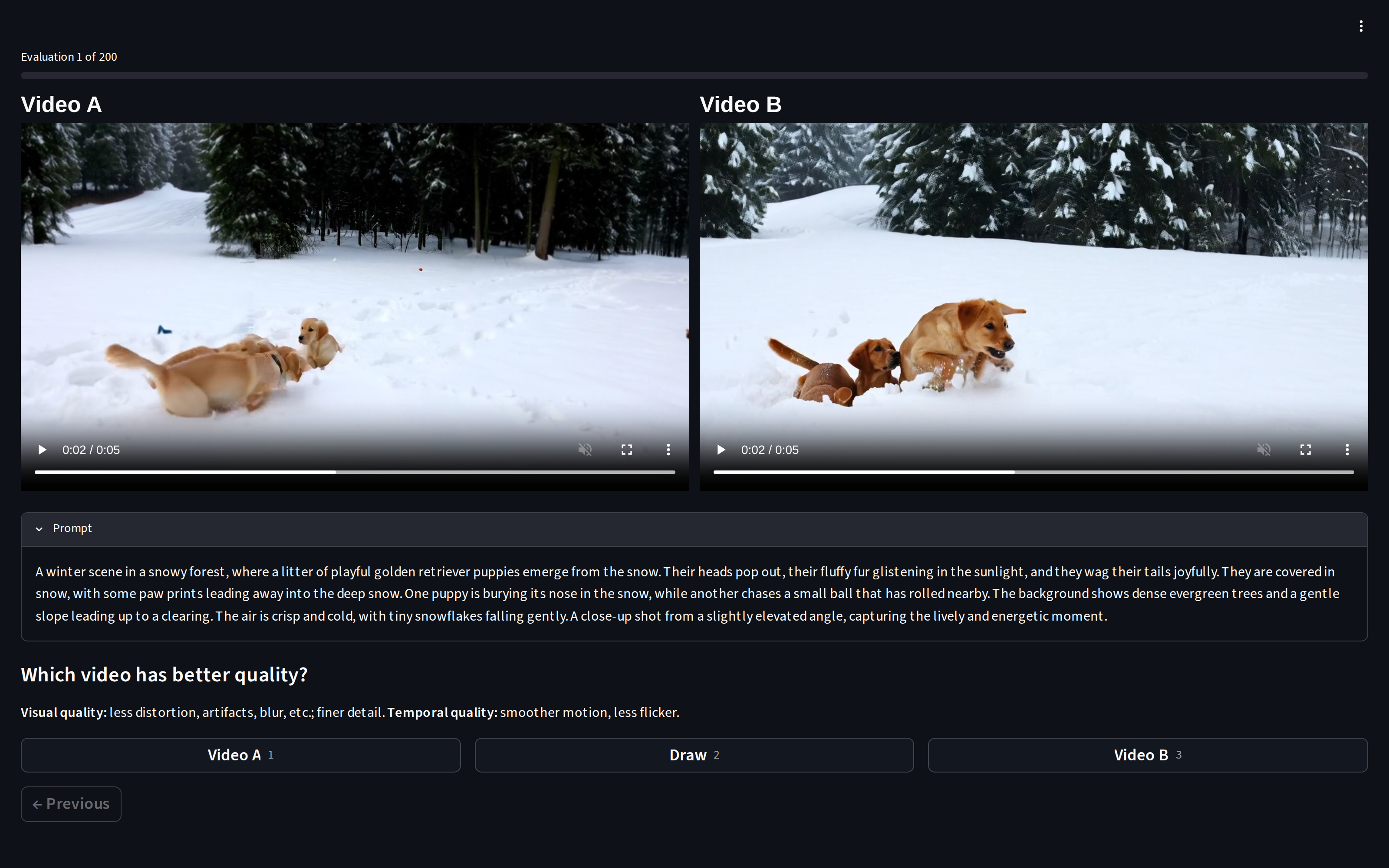}
  \caption{\textbf{User study interface.} The two videos play side by side; the vote buttons are below. The prompt box
    is opened here for illustration; raters see it collapsed.}
  \label{fig:human_ui}
\end{figure}

We run a blind pairwise study: in each pair, raters watch two videos generated from the same prompt, one by our method
and one by a baseline, and pick the better one or a draw.

\paragraph{Interface.} The two videos play side by side, in sync and looped (\cref{fig:human_ui}). LR videos
($640{\times}352$) are upscaled to $1280{\times}704$ by the browser. Raters are told to judge visual quality
(artifacts, blur, detail) and temporal quality (smooth motion, no flicker), and to choose Draw only if they cannot tell
which is better.

\paragraph{Prompts and pairs.} We randomly choose 100 prompts from MovieGen and generate one video per method with the
same seed, giving 400 pairs (ours vs.\ each of the 4 baselines). The videos are uncurated. Each of the 15 raters judges
200 pairs, shown in random order with ours randomly on the left or right.

\paragraph{Win rate.} The win rate is (ours better $+$ draw$/2$) / votes~\cite{alpaca_eval,genai_arena}. \Cref{tab:human}
lists the raw vote shares behind the win rates.

\begin{table}[h!]
\centering
\caption{\textbf{Human evaluation} on Wan2.1 14B ($k=25$), 750 votes per baseline (749 for Foveated Diffusion).
``Win rate'' counts a draw as half a vote for each side.}
\label{tab:human}
\small
\setlength{\tabcolsep}{3pt}
\resizebox{\columnwidth}{!}{%
\begin{tabular}{@{}lcccc@{}}
\toprule
Ours vs. & Ours better & Draw & Baseline better & Win rate \\
\midrule
HR                 & 23.2\% & 44.0\% & 32.8\% & 45.2\% \\
LR                 & 51.7\% & 34.7\% & 13.6\% & 69.1\% \\
Phase-Aligned RoPE & 97.9\% & 1.9\%  & 0.3\%  & 98.8\% \\
Foveated Diffusion & 84.6\% & 10.8\% & 4.5\%  & 90.1\% \\
\bottomrule
\end{tabular}
}
\end{table}

\section{Dual-resolution attention ablation details}\label{sec:supp_kv_ablation}

This section gives the details of the dual-resolution attention ablation in \cref{sec:ablation}. We train two Wan2.1
1.3B (480p) models in the same way for 5,000 steps. One uses dual-resolution attention (ours). The other uses a single
shared K/V with adjusted RoPE, as in Foveated Diffusion~\cite{foveateddiffusion} and Phase-Aligned
RoPE~\cite{wu2025phase}. To test only the attention, both models use the same mixed-resolution decoder and the same
moving circular HR region. Each model generates one video for each of the 1,003
MovieGen prompts, with the same seeds. In each video, we take 80 boxes of $64{\times}64$ pixels centered on the HR/LR
edge. Temporal LPIPS~\cite{chu2020tecogan} measures how much a box changes between consecutive frames. The warping
error~\cite{lai2018blind} measures the change that remains after removing motion with optical flow
(RAFT~\cite{teed2020raft}). Ours lowers temporal LPIPS from $4.59$ to $4.14 \times 10^{-2}$ ($10\%$ lower) and the
warping error from $2.61$ to $1.96 \times 10^{-3}$ ($25\%$ lower) (\cref{tab:kv_ablation}). So content at the seam
distorts less from frame to frame: objects stay consistent as the edge moves over them, instead of distorting.

\begin{table}[h]
\centering
\caption{\textbf{Temporal consistency at the HR/LR edge} ($64{\times}64$ boxes centered on the edge, 1,003 MovieGen
prompts, Wan2.1 1.3B). \textbf{Bold}: better of the two K/V designs.}
\label{tab:kv_ablation}
\small
\setlength{\tabcolsep}{3pt}
\begin{tabular}{@{}lcc@{}}
\toprule
Metric & Shared K/V & Dual K/V (ours) \\
\midrule
Temporal LPIPS ($\times 10^{-2}$) $\downarrow$ & 4.59 & \textbf{4.14} \\
Warping error ($\times 10^{-3}$) $\downarrow$  & 2.61 & \textbf{1.96} \\
\bottomrule
\end{tabular}
\end{table}

\section{LPIPS label details}\label{sec:supp_lpips}

\begin{figure}[h!]
  \centering
  \includegraphics[width=\columnwidth]{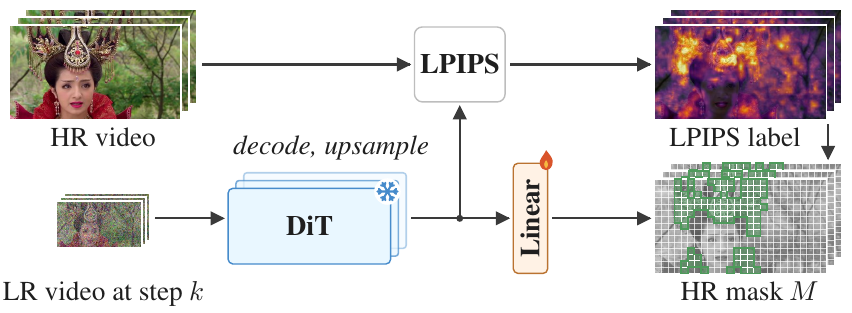}
  \caption{\textbf{LPIPS label}, an end-to-end alternative to the subject label in \cref{fig:label_pipeline}. \emph{Top:}
    LPIPS between the real HR video and the DiT's decoded, upsampled LR prediction. \emph{Bottom:} a linear head on
    the frozen DiT learns to predict it.}
  \label{fig:lpips_label}
\end{figure}

This section details the perceptual label of \cref{sec:region} (\cref{fig:lpips_label}), which directly uses the
DiT's own denoising error. We noise the 240p latent of
a real video to step $k$, and let the frozen DiT predict the clean video $\hat{x}_0$. Each token's label is the
LPIPS~\cite{zhang2018lpips} between the decoded, $2\times$-upsampled $\hat{x}_0$ and the real 480p video $x^{\mathrm{HR}}$,
averaged over the token:
\begin{equation}
  L = \mathrm{LPIPS}\big(\mathrm{Up}(\mathrm{Dec}(\hat{x}_0)),\, x^{\mathrm{HR}}\big).
  \label{eq:lpips_label}
\end{equation}
From the DiT's middle-block feature at step $k$, the linear head learns to predict $L$ for each denoised sample, with a
regression loss and a ranking loss within each frame. Importantly, this prediction also serves as a confidence score
for each denoised patch: a higher score means the model expects a larger difference from the ground truth. We train it
on real videos, with 8 noise draws per video, for
3,000 iterations at batch size 8 on 8 GPUs. At inference, $M$ is the $25\%$ of tokens with the highest predicted $L$
in each frame.

\section{Mixed-resolution decoder training recipe}\label{sec:supp_mixer_loss}

We train the mixer $G$ and upscaler $\mathcal{U}$ on real clips. For each training clip, we encode the HR video and its
$2\times$-downsampled LR video with the VAE encoder to obtain the ground-truth HR latent $z^h_0$ and LR latent $z^l_0$.
To form a training pair that matches the output of mixed-resolution denoising, we take a training mask $M$
(\cref{sec:method}) and keep only the HR latent inside $M$ ($z^h = M_{\uparrow}\odot z^h_0$) and only the LR latent
outside it ($z^l = (1-M)\odot z^l_0$). Given $(z^h, z^l, M)$, we train the merged latent $\hat{z}$ (\cref{eq:mix_out})
to predict the full ground-truth HR latent $z^h_0$. We use the loss from~\cite{wu2025phase} for the upscaler training,
which combines a latent, a pixel and a perceptual term:
\begin{equation}
  \begin{aligned}
    \mathcal{L}_{\mathrm{mix}} &= \|\hat{z}-z^h_0\|_1+\|\mathrm{Dec}(\hat{z})-\mathrm{Dec}(z^h_0)\|_1\\
      &+ \lambda\,\mathrm{LPIPS}\big(\mathrm{Dec}(\hat{z}),\mathrm{Dec}(z^h_0)\big),
  \end{aligned}
  \label{eq:mix_loss}
\end{equation}
where $\mathrm{Dec}$ is the VAE decoder and $\lambda$ a weight. In practice, the latent term is computed only outside
the HR region $M_{\uparrow}$, and both L1 terms are up-weighted near the boundary. We train on 30,000 real video clips
for 4,000 iterations at batch size 64.

\section{GPT-5.6 pairwise test details}\label{sec:supp_gpt}

GPT-5.6 sees two anonymous images (A and B) from the same prompt and seed, one ours and one from a baseline. The
prompt is hidden, so it rates image quality only: fine detail, seams, blocks, smearing, distorted objects and
garbled text. It may zoom in but may not compute metrics. Each pair is judged once, our image is A in half of
the pairs, and ties were allowed but never chosen.

\end{document}